\pdfoutput=1
\documentclass{article}

\usepackage{microtype}
\usepackage{graphicx}
\usepackage{subfigure}
\usepackage{booktabs} % for professional tables
\usepackage[draft, cachedir=.]{minted} 
\usepackage{listings}
\usepackage{enumitem}
\usepackage{amsmath}
\usepackage{amssymb}
\usepackage{wrapfig}

\usepackage[final]{corl_2020} % Uncomment for the camera-ready ``final'' version.

\title{Interactive Visualization for Debugging RL}

\author{
Shuby Deshpande, \quad Benjamin Eysenbach, \quad Jeff Schneider \\
School of Computer Science, \\
Carnegie Mellon University, \\
Pittsburgh, PA 15213. \\
\texttt{\{shubhand,beysenba,schneide\}@cs.cmu.edu}
}
\begin{document}
\maketitle

%===============================================================================

\begin{abstract}
Visualization tools for \emph{supervised learning} allow users to interpret, introspect, and gain an intuition for the successes and failures of their models.
%%BE.7.23: Add a citation. Generally, writing in the present tense.
While \emph{reinforcement learning} practitioners ask many of the same questions, existing tools are not applicable to the RL setting as these tools address challenges typically found in the supervised learning regime.
%%BE.7.23: Explain why.
In this work, we design and implement an interactive visualization tool for debugging and interpreting RL algorithms. Our system\footnote{A functional interactive demo of the system can be at \url{https://tinyurl.com/y5gv5t4m}} addresses many features missing from previous tools such as (1) tools for supervised learning often are not interactive; (2) while debugging RL policies researchers use state representations that are different from those seen by the agent; (3) a framework designed to make the debugging RL policies more conducive. We provide an example workflow of how this system could be used, along with ideas for future extensions.
\end{abstract}

% Two or three meaningful keywords should be added here
\keywords{Reinforcement Learning, Interpretability, Visualization} 

%===============================================================================

\section{Introduction}
\label{sec:intro}

Machine learning systems have made impressive advances due to their ability to learn high dimensional models from large amounts of data \citep{lecun_deep_2015}. However,  high dimensional models are hard to understand and trust \citep{doshi-velez_towards_2017}. Visualization systems are important for overcoming these challenges. Many tools exist for addressing these challenges in the \emph{supervised learning} setting, which find usage in tracking metrics \citep{satyanarayan_vega-lite_2017}, generating graphs of model internals \citep{wongsuphasawat_visualizing_2018}, and visualizing embeddings \citep{van_der_maaten_visualizing_2008}. However, there is no corresponding set of tools for the reinforcement learning setting. At first glance, it appears we may repurpose existing tools for this task. However, we quickly run into limitations, which arise due to the intent with which these tools were designed in the first place. Reinforcement learning (RL) is a more interactive science \citep{neftci_reinforcement_2019} compared to supervised learning, due to a stronger feedback loop between the researcher and the agent. Whereas supervised learning involves a static dataset, RL entails collecting new data. To fully understand an RL algorithm, we must understand the effect of the RL algorithm on the data collected. Note that in supervised learning, the learned model has no effect on the fixed dataset."

Visualization systems at their core consist of two components: \emph{representation} and \emph{interaction}. \emph{Representation} is concerned with how data is mapped to a representation and then rendered. \emph{Interaction} is concerned with the dialog between the user and the system as the user explores the data to uncover insights \cite{yi_toward_2007}. These may appear to be disparate, but in actuality, it is hard to discount the influence that each have on each other. The tools we use for representation affect how we interact with the system, and our interaction affects the representations that we create. Thus, while designing visualization systems, it is important to think about the intended application domain, in this case, reinforcement learning.

% One useful framework to think about when designing features that our visualization systems should encapsulate, is the human action cycle, as introduced in \citep{norman_design_2001}. After these features have been designed, they must be evaluated to ensure they satisfy the original goals.

Three dimensions along which to evaluate visualization systems, as proposed by \citep{beaudouin-lafon_designing_2004}, and adapted here for relevance, are:
\begin{itemize}[noitemsep, nolistsep]
    \item[-] \textit{descriptive power}: the ability to describe a significant range of existing interfaces
    \item[-] \textit{evaluative power}: the ability to help assess multiple alternatives
    \item[-] \textit{generative power}: the ability to help create new designs
\end{itemize}

Existing tools primarily focus on descriptive power. Using them, we can plot common descriptive metrics such as cumulative reward, TD-error, and action values, to name a few. However, these systems either lack or are deficient in evaluative and generative power. Ideally, the systems we use should help us answer questions such as:

\nocite{noauthor_tensorboard_2020}

\begin{itemize}[noitemsep, nolistsep]
    \item[-] What sequence of dynamics cause the resulting agent behavior?
    \item[-] What effects do noteworthy states have on the resulting policy? Are there other states which lead to similar outcomes?
    \item[-] How does the agent state visitation distribution change as training progresses?
    % \item[-] If the current iteration of the agent had not experienced this state, would that have resulted in a significant policy change (i.e. the counterfactual)?
    % \item[-] How does the agent state visitation distribution change in the context of policy updates?
    % \item[-] What actions should I take to induce the intended agent behavior?
\end{itemize}

These are far from an exhaustive list of questions that a researcher may pose while training agent policies, but are chosen to illustrate the constraints that our current set of tools impose against being able to easily answer questions of this nature.

This paper describes our attempt at constructing Vizarel \footnote{Vizarel is a portmanteau of visualization + reinforcement learning.}, an interactive visualization system to help debug and interpret RL algorithms, that can answer these questions. Towards these goals, we identify features which an interactive system for interpretable reinforcement learning should~encapsulate, and build a prototype of these ideas. Further, we describe a guiding framework around which future additions could be built. We complement this by providing a walkthrough example of how this system can fit into the RL workflow and used in a real scenario to debug a policy.

%===============================================================================

\section{Related Work}
\label{sec:related_work}

%%BE.7.28: Check below.
Existing visualization tools for machine learning focus on the supervised learning setting. As we have argued in the introduction, the process of designing and debugging RL algorithms requires a different set of tools. In the rest of this section, we highlight aspects of prior work upon which our system builds.
% To the best of our knowledge, there do not exist visualization systems built for interpretable reinforcement learning that effectively address the broader goals we have identified. There exists prior work aspects of which are relevant to features which the current system encapsulates, which we now detail.

\textbf{Visual Interpretability}
Related work for increasing understanding in machine learning models using visual explanations includes
\cite{olah_feature_2017, simonyan_deep_2014, zeiler_visualizing_2013} focus on feature visualization in neural networks, \cite{strobelt_lstmvis_2017, kahng_activis_2017, kapoor_interactive_2010, krause_interacting_2016, yosinski_understanding_nodate} present visual analysis tools for different variants of machine learning models, and \cite{olah_building_2018} which treats existing methods as composable building blocks for user interfaces.

\textbf{Explaining agent behavior}
There exists past work that tries to explain agent behavior. \cite{amir_highlights_nodate} summarizes agent behavior by displaying important trajectories. \cite{van_der_waa_contrastive_2018} introduces a method to provide contrastive explanations between user derived and agent learned policies. \cite{huang_enabling_2017} details showing maximally informative examples to guide the user towards understanding the agent objective function. \cite{hayes_improving_2017} present algorithms and a system that enables robots to synthesize policy descriptions and respond to human queries. This work is similar to \cite{amir_highlights_nodate, van_der_waa_contrastive_2018, huang_enabling_2017}, in there being a motivation to provide relevant information to the researcher to easily explore the possible space of solutions while debugging the policy. Similar to \cite{hayes_improving_2017}, we present a functioning system which can respond to human queries to provide explanations. However, in contrast, the interactive system we present is built around the RL training workflow, and designed to evolve beyond the explanatory use case to serve as a supplement to the existing ecosystem of widely used tools \cite{noauthor_tensorboard_2020, satyanarayan_vega-lite_2017}.

%% deciding where (and if) to insert this. %%
%% perhaps -> "initial discussions with the interpretable machine learning community resulted in valuable feedback and focusing of efforts on building relevant features."

\nocite{deshpande_vizarel_2020}

%===============================================================================

%% removing this figure for conciseness. concepts covered through other figures %%

\section{Preliminaries}
\label{sec:preliminaries}
We use the standard reinforcement learning setup~\citep{sutton_reinforcement_2018}. An agent interacting with an environment at discrete timesteps $t$, receiving a scalar reward $r(s_t, a_t) \in \mathbb{R}$.
The agent's behavior is defined by a policy $\pi$, which maps states $s \in S$, to a probability distribution over actions, $\pi: S \rightarrow P(A)$. The environment can be stochastic, which is modeled by a Markov decision process with a state space $S$, action space $A \in R^n$, an initial state distribution $p(s_0)$, a transition function $p(s_{t+1} \mid s_{t}, a)$, and a reward function $r(s_t, a_t)$. The future discounted return from a state $s_t$ and action $a_t$ is defined as $R_t = \sum_{i=t}^{T} \gamma^{i-t}r(s_t,a_t)$, with a discount factor $\gamma \in [0,1]$.
We use a replay buffer \citep{mnih_playing_2013, lin_self-improving_nodate} to store the agent's experiences $e_t = (s_t, a_t, r_t, s_{t+1})$ in a buffer $B = \{e_0, e_1, ..., e_T\}$.

%===============================================================================

\section{\textbf{Vizarel}: A Tool for Interactive Visualization of RL}
\label{sec:system_desc}
This section will describe how our interactive visualization system (Vizarel), is currently designed. The system offers different views which encapsulate the spatial and temporal dimensions of agent policies. The tool consists of a set of ``viewport'' modules. Each viewport is an abstract entity that can be backed by different \textit{specs}, conditioned on the underlying data stream. Specs are fundamental visualization elements that can be combined in different ways to generate viewports, examples of which include:
\begin{enumerate}[noitemsep, nolistsep]
    \item \textit{image buffers}: to visualize image based observation spaces (or non-image based spaces if rendered)
    \item \textit{line plots}: to visualize data ordered by a sequential dimension, such as action values or rewards across time
    \item \textit{scatter plots}: to visualize embedding spaces \cite{van_der_maaten_visualizing_2008} or compare tensors along specified dimensions
    \item \textit{histograms}: to visualize frequency counts of specified tensors or probability distributions
\end{enumerate}
Viewports can be combined in various ways to create different views on the underlying data stream. This naturally leads to the idea of plugins that can be can be integrated into the core system to support different visualization schemes and algorithms. For example, the user could combine image buffers and line plots in novel ways to create a viewport for the state-action value function \cite{sutton_reinforcement_2018}. This viewport could then be released as a plugin and distributed for use amongst a broader base of users. In the rest of this section, we describe two types of some of the viewports currently implemented in Vizarel: temporal viewports and spatial viewports. 
% Section~\ref{sec:XX} will describe how new viewports can be readily added to Vizarel.

\subsection{Temporal Views}
\label{subsec:temporal_views}

Temporal views are oriented around visualizing the data stream (e.g. images, actions, rewards) as an ordered sequence of events. We have implemented three types of temporal viewports: state viewports, action viewports, and reward viewports.

\subsubsection{State Viewport}
\label{subsubsec:state_viewport}

\begin{figure}[t]
    \centering
    \includegraphics[width=\linewidth]{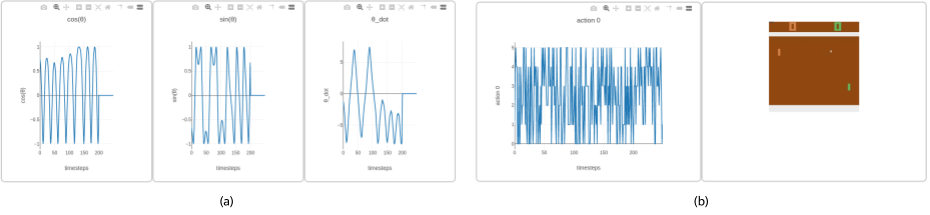}
    \caption{\textbf{State + Action Viewports}
    \small
    \textbf{(a)} Visualizing the state viewport for the inverted pendulum task. This representation overlayed with another state viewport similar to (b), provides the user with better intuition about the correspondence between states and actions for non image state spaces, which is easier to interpret. \textbf{(b)} Visualizing the action viewport for the Pong environment \cite{bellemare_arcade_2013}. Hovering over instantaneous timesteps dynamically updates the state viewport (\ref{subsubsec:state_viewport}) and shows the corresponding rendered image for the selected state. This representation, provides the user better intuition about the current agent policy, and could help subsequent debugging.}
    \label{fig:sa_viewport}
    \vspace{-1em}
\end{figure}

Referring to the state-space formulation from \S \ref{sec:preliminaries}, states can primarily be classified as either image-based or non-image based spaces. The type of observation space influences the corresponding spec which is used to generate the viewport. We provide two examples that illustrate how these differing specs can result in different viewports. Consider a non-image based observation space, such as that for the inverted pendulum task. Here, the state vector $\vec{s} = \{\sin(\theta), \cos(\theta), \dot{\theta}\}$, where $\theta$ is the angle which the pendulum makes with the vertical.

\nocite{brockman_openai_2016}

We can visualize the state vector components individually, which provides insight into how states vary across episode timesteps (Figure \ref{fig:sa_viewport}). Since an image representation is easier for humans to interpret, we can generate an additional viewport backed by an image buffer spec, which tracks the corresponding changes in image space. Having this simultaneous visualization is useful since this now enables us to jump back and forth between the state representation which the agent receives, and the corresponding element in image space, by simply hovering over the desired timestep in the state viewport.

For environments that have higher dimensional non-image states, such as that of a robotic arm with multiple degrees of freedom, we visualize individual state components. However, since this may not be intuitive, we also generate an additional viewport as an overlay to display an image rendering of the environment.

\subsubsection{Action Viewport}
\label{subsubsec:action_viewport}

\begin{figure}[t]
    \centering
    \includegraphics[width=0.8\linewidth]{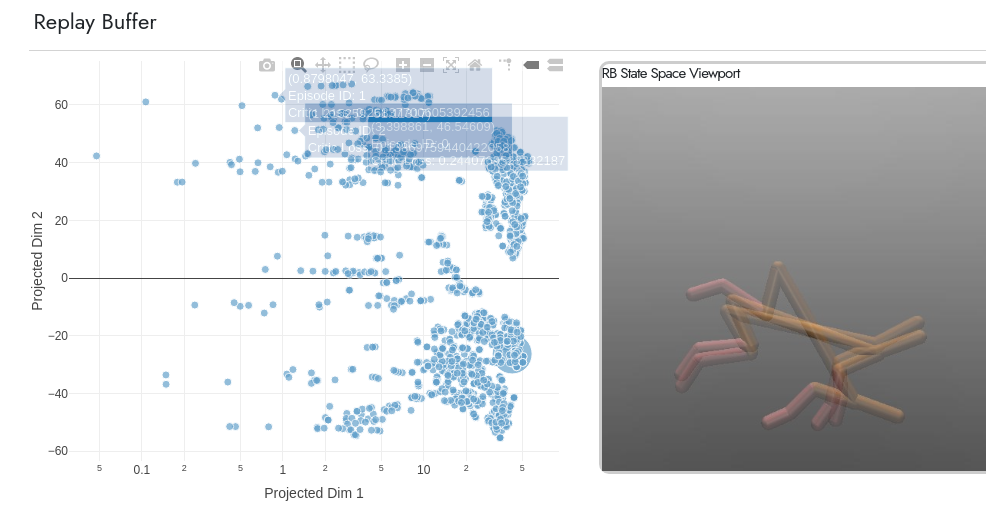}
    \caption{\textbf{Replay Buffer Viewport}
    \small
    Projecting the contents of the replay buffer into a 2D space for easier navigation. This provides a proxy to the replay buffer diversity and can help in subsequent debugging. Hovering over points in the replay buffer dynamically updates the generated state viewport (\ref{subsubsec:state_viewport}), and shows the rendered image for the corresponding state (animation depicted using overlay).
    }
    \label{fig:rb_viewport}
    \vspace{-1em}
\end{figure}

The action viewport visualizes how the action $a_t$ varies across the episode by creating a viewport backed by a line plot spec (Figure \ref{fig:sa_viewport}). For agents where we have access to a distribution over actions instead, we can generate a viewport backed by a histogram spec, and visualize how the action distribution changes over time. A similar technique could be used to visualize how the state-action value distribution changes over time for groups of similar states.

\subsubsection{Reward Viewport}
\label{subsubsec:reward_viewport}

The reward is typically a scalar quantity, which motivates generating a viewport backed by a line plot spec. A user can look at the reward viewport together with the state-space viewport to understand which states result in high reward.
For most agent environments, the reward function comprises of different components weighted by different coefficients. These individual components are often easier to interpret since they are usually backed by a physically motivated quantity tied to agent behaviors that we wish to either reward or penalize. In situations where we have access to these reward components, we can generate multiple viewports each of which visualizes different components of this reward function~vector. Such a visualization could help the user design reward functions that are immune to reward hacking \cite{amodei_concrete_2016}, by providing the user more insight into the correspondence between states, actions, and the components of the reward function which the agent is attempting to maximize.

\subsection{Spatial Views}
\label{subsec:spatial_views}

Spatial views are oriented around visualizing the data stream as a spatial distribution of events. We have implemented three types of spatial viewports: replay buffer viewports, distribution viewports, tensor comparison viewports, and trajectory viewports.

\subsubsection{Replay Buffer Viewport}
\label{subsubsec:rb_viewport}

As formulated in \S \ref{sec:preliminaries}, the replay buffer stores the agent's experiences $e_t = (s_t, a_t, r_t, s_{t+1})$ in a buffer $B = \{e_0, e_1, ..., e_T\} \forall i \in [0, T]$.
For off-policy algorithms, the replay buffer is of crucial importance, since effectively serves as an online dataset for agent policy updates \cite{fu_d4rl_2020}. For visualizing datasets, there exist tools \cite{noauthor_facets_2020}, which provide the user with an intuition for the underlying data distribution. Similar intuitions can be provided by visualizing the space of points in the replay buffer.

Since the individual elements of the replay buffer are at least a four-dimensional vector $e_t$, this rules out the possibility of generating viewports backed by specs in the original space. We can instead visualize the replay buffer samples by transforming the points \citep{van_der_maaten_visualizing_2008} to a lower-dimensional representation. This technique provides insight into the distribution of points in the replay buffer, which leads to a visualization of the replay buffer diversity \cite{de_bruin_experience_nodate}.

The size of the replay buffer can be quite large \cite{zhang_deeper_2018}, which could lead to difficulties while navigating the space of points visualized in the replay buffer viewport. To nudge \cite{thaler_nudge_2009} the user behavior towards investigating samples of higher potential interest, we scale the size of points in proportion to the absolute normalized TD error, which has been used in past work \cite{schaul_prioritized_2016} as a measure of sample priority during experience replay.

The replay buffer viewport can naturally be combined with the state viewport, to simultaneously visualize the image representation of the state, by tracking changes as the user hovers over points in the replay buffer viewport (Figure \ref{fig:rb_viewport}).

\subsubsection{Distribution Viewport}
\label{subsubsec:dist_viewport}

\begin{figure}[t]
    \centering
    \includegraphics[width=0.8\linewidth]{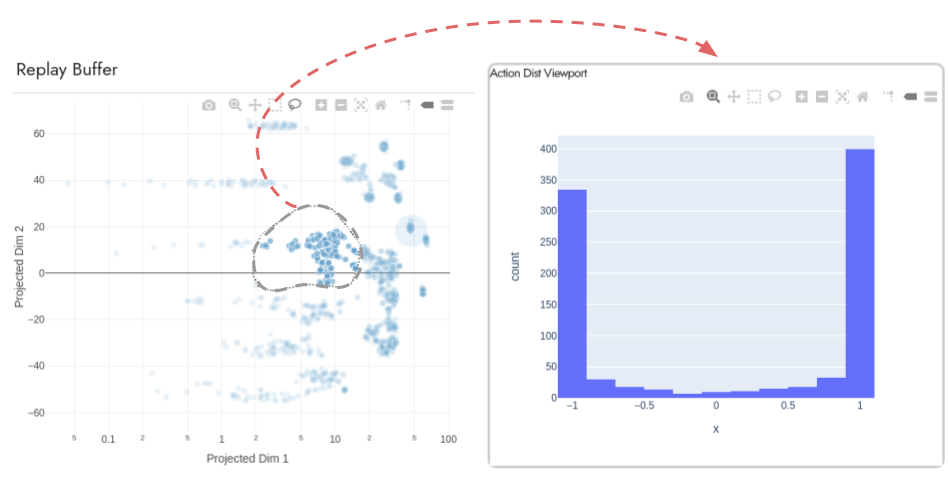}
    \caption{\textbf{Distribution Viewport}
    \small
    Using the lasso tool to select a group of points (dashed gray line) in the replay buffer viewport (\ref{subsubsec:rb_viewport}), dynamically updates (dashed red line) the distribution viewport (\ref{subsubsec:dist_viewport}) by computing and plotting the distribution of values for the specified tensor (e.g. actions or rewards).
    }
    \label{fig:dist_viewport}
    \vspace{-1em}
\end{figure}

The distribution viewport (Figure \ref{fig:dist_viewport}) complements the replay buffer viewport by allowing the user to ask questions concerning the distribution of action, rewards, or other relevant data streams for selected groups of points.
% \begin{itemize}
%     \item What is the distribution of actions for the selected group of points?
%     \item What is the distribution of rewards for the selected group of points?
% \end{itemize}

If policy updates result in task learning, intuition suggests the entropy of the action distribution should reduce over time (discounting any external annealing caused due to exploration), which can be easily verified through this viewport. In the limit, the distribution of actions for a group of similar points should converge to a Dirac distribution, since the optimal policy for an infinite horizon discounted MDP is a stationary distribution \cite{puterman_markov_2005}. In practice, observing the distribution converging around the mean value could indicate a promising policy training experiment.

For multi-dimensional action spaces, the viewport could be repurposed to display the variance of the action distribution, plot different slices of the action distribution, or use more sophisticated techniques~\cite{huber_projection_1985}.

\begin{figure}[t]
    \centering
    \includegraphics[width=0.8\linewidth]{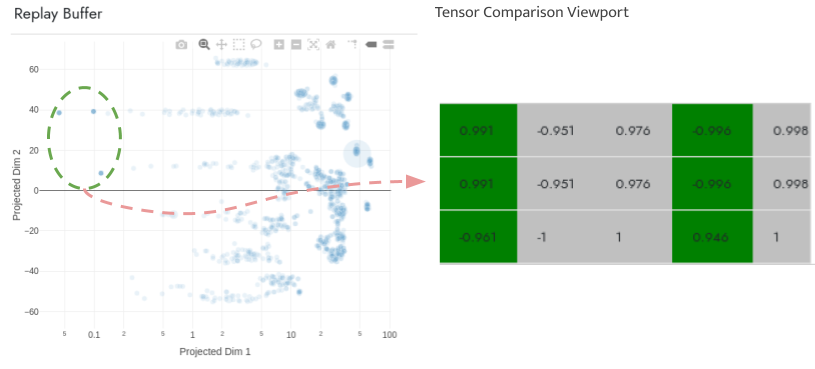}
    \caption{\textbf{Tensor Comparison Viewport}
    \small
    Selecting points (dashed green line) in the replay buffer viewport (\ref{subsubsec:rb_viewport}), and generating (dashed red line) the tensor comparison viewport (\ref{subsubsec:t_compare_viewport}), allows the user to compare the specified tensors (e.g. actions or states), where dimensions of higher variance are automatically highlighted. This could lead to faster debugging in environments where each dimension corresponds to physically intuitive quantities. 
    \label{fig:tensor_compare}}
    \vspace{-1em}
\end{figure}

\subsubsection{Tensor Comparison Viewport}
\label{subsubsec:t_compare_viewport}

For environments that have higher dimensional action spaces, it is hard for the user to understand how neighboring points in the replay buffer viewport differ. This becomes especially relevant for diagnosing clusters of points that have a higher TD error. The tensor comparison viewport (Figure \ref{fig:tensor_compare}) enables the user to easily select points and then compare them along the dimensions of interest, which for example could be actions. Dimensions that have a standard deviation beyond a specified threshold are automatically highlighted, which enables the user to focus on the dimensions of interest.

\begin{figure}[t]
    \centering
    \includegraphics[width=0.7\linewidth]{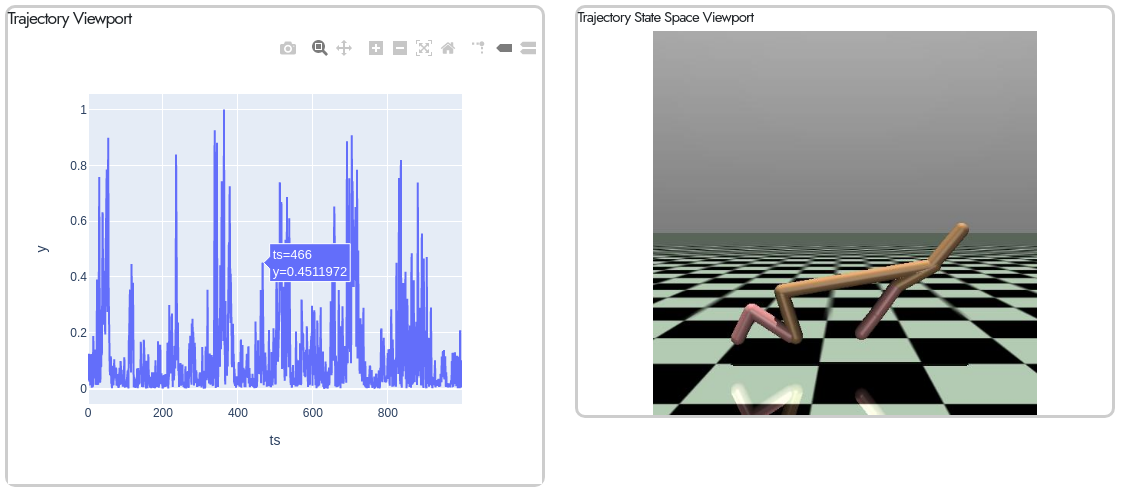}
    \caption{\textbf{Trajectory Viewport}
    \small
    Selecting points in the replay buffer viewport (\ref{subsubsec:rb_viewport}), causes the trajectory viewport (\ref{subsubsec:trajectory_viewport}) to dynamically update and plot the absolute normalized TD error values over the length of the trajectory. Hovering over points in the trajectory viewport, allows the user to view a rendering of the state corresponding to that timestep in the generated state viewport (\ref{subsubsec:state_viewport}).
    }
    \label{fig:t_viewport}
    \vspace{-1em}
\end{figure}

\subsubsection{Trajectory Viewport}
\label{subsubsec:trajectory_viewport}

A fusion of the components from the spatial and temporal views leads to the notion of a \textit{spatio-temporal view}, such as that of the trajectory viewport (Figure \ref{fig:t_viewport}). The replay buffer viewport by itself visualizes the spatial nature of the points in the replay buffer but does not display the temporal nature of trajectories. Being able to switch between spatial and temporal views is crucial when understanding and debugging policies. This is supported by selecting points in the replay buffer viewport, which then retrieves the corresponding trajectory.

The trajectory viewport is backed by a line plot spec, where the X coordinate represents the timestep and the Y coordinate is the absolute TD error, normalized to lie within $[0, 1]$. Hovering over points in the trajectory viewport retrieves a rendering of the corresponding state in an instantiation of the state viewport. This correspondence enables the user to easily navigate to, and visualize, snippets in the trajectory that have a high TD error, thus speeding up debugging of policies.

%===============================================================================

\section{Walkthrough}
\label{sec:walkthrough}

\begin{figure}[t]
    \centering
    \includegraphics[width=0.8\linewidth]{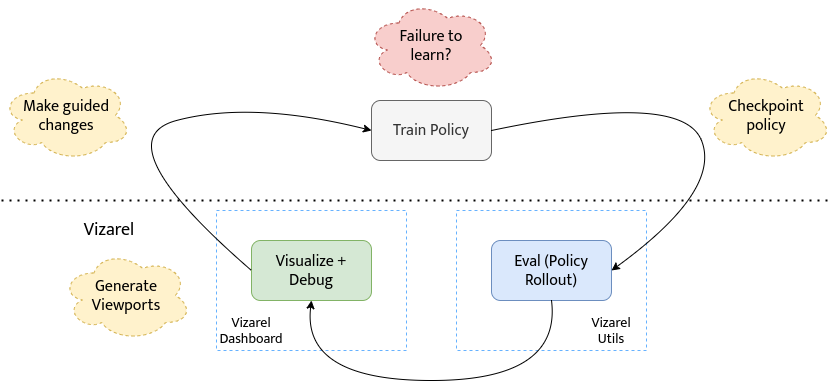}
    \caption{\textbf{Vizarel Workflow Diagram}
    \small
    Typical steps during policy debugging, and how the designed system fits into this workflow. The system takes as input a policy saved during a checkpoint and evaluates the policy through a specified number of rollouts. This data is then visualized through viewports which the user specifies, and used for debugging the policy through making guided changes.
    }
    \label{fig:vizarel_usage}
    \vspace{-1em}
\end{figure}

% \begin{figure}[t]
\begin{wrapfigure}{R}{0.5\textwidth}
    \centering
    \includegraphics[width=\linewidth]{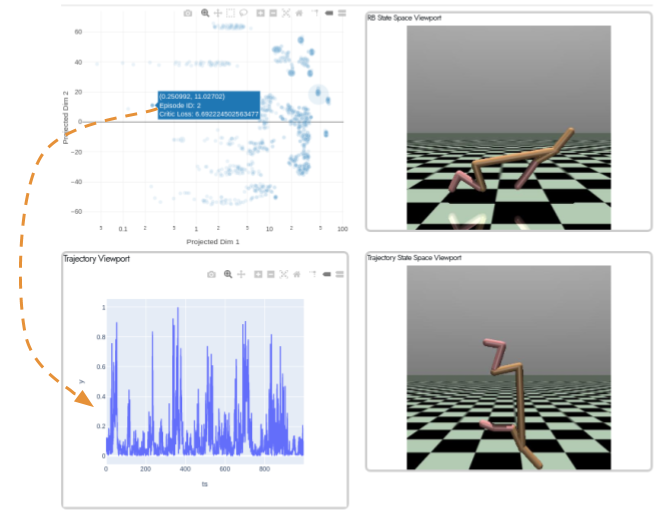}
    \caption{\textbf{Spatio-Temporal Interaction}
    \small
    Visualizing the replay buffer viewport (\ref{subsubsec:rb_viewport}) (spatial view), and trajectory viewport (\ref{subsubsec:trajectory_viewport}) (temporal view), along with overlays to independently track image of states in both as a state space viewport (\ref{subsubsec:state_viewport}). Navigating between these viewports allows the user to observe agent behavior in along both spatial and temporal dimensions, which could facilitate better insights during debugging.
    }
    \label{fig:spatio_temporal}
    \vspace{-1em}
% \end{figure}
\end{wrapfigure}
We now detail an example workflow of how the system can be used in a real scenario. Figure \ref{fig:vizarel_usage}, illustrates how Vizarel fits into the typical sequence of steps in an RL researcher's policy debugging workflow. Training a successful agent policy often requires multiple iterations of changing algorithm hyperparameters, and changing design decisions.

To speed up and increase intuition in this process, the researcher can load a stored checkpoint of the policy into the system, and evaluate a specified number of policy rollouts. Empirically, we've found that there should be enough rollouts to ensure sufficient coverage of the state space, as this influences the scope of insight during downstream debugging. These rollouts can then be visualized and interacted with through specifying the required data streams and generating different viewports.

\begin{figure}[t]
    \centering
    \includegraphics[width=0.7\linewidth]{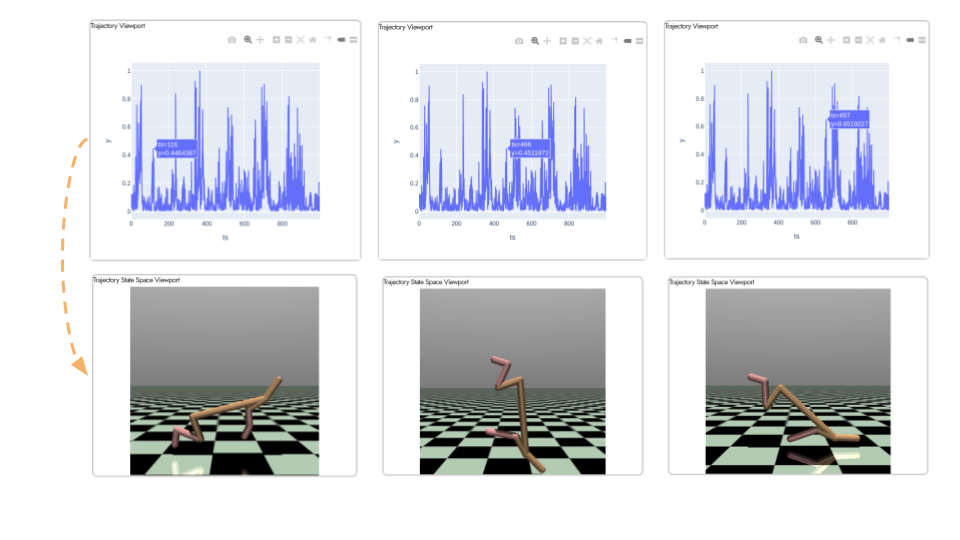}
    \caption{\textbf{Comparing TD error along an agent trajectory}
    \small
    Visualizing the trajectory viewport (\ref{subsubsec:trajectory_viewport}), allows the user to compare the TD error at different timesteps along the trajectory, along with the associated state viewport (\ref{subsubsec:state_viewport}). An example interaction is visualized here by hovering over regions of potential interest in the trajectory viewport. This simultaneous view allows the user to easily compare and draw similarities between action sequences which cause large changes in TD error.
    \label{fig:td_error}}
    \vspace{-1em}
\end{figure}

Figure \ref{fig:spatio_temporal}, shows an example of replay buffer, state, and trajectory viewports generated for a policy trained using DDPG on the HalfCheetah task. The high variance in the TD error suggests the presence of critic overestimation bias \cite{thrun_issues_nodate}, which could be remedied by using algorithms known to reduce the impact of this issue \cite{fujimoto_addressing_2018, hasselt_double_2010}. Figure \ref{fig:td_error} shows how the user can compare the TD error along the agent trajectory. Hovering over regions of potential interest in the trajectory viewport allows the user to find action sequences that cause high variance in TD error. A similar technique could be used to visualize clusters of states in the replay buffer space with high TD error (Figure \ref{fig:rb_viewport}).
This approach could enable the user to identify patterns in states across space or time that persistently have high TD error, and design methods to mitigate this \cite{amodei_concrete_2016}.

Another approach the user could take is to generate a distribution viewport (Figure \ref{fig:dist_viewport}), and identify the distribution of actions in the vicinity of states with a high TD error. If similar states persistently have a higher action and/or reward variance, this suggests that the usage of variance reduction techniques could help learning \cite{schulman_high-dimensional_2018, romoff_reward_2018}.
Once promising avenues for modification have been identified, the user can make guided changes, and retrain the policy.

%===============================================================================

\section{Conclusion}
\label{sec:conclusion}

In this paper, we have introduced a visualization tool, Vizarel, that helps interpret and debug RL algorithms. Existing tools which we use to gain insights into our agent policies and RL algorithms are constrained by design choices that were made for the supervised learning framework. To that end, we identified features that an interactive system for debugging and interpreting RL algorithms should encapsulate, described a guiding framework for system design and implementation, and provided a walkthrough of an example workflow the user could follow to gain insights into a trained agent policy using this tool.

There are multiple features under development that contribute towards both the core system. One feature is the integration of additional data streams such as saliency maps \citep{greydanus_visualizing_2018} to complement the state viewport. Another is designing the capability to use the system in domains that lack a visual component (e.g. healthcare \cite{yu_reinforcement_2020} and education \cite{reddy_accelerating_nodate}). An extension is to add search capabilities that allow the user to easily traverse, query, and identify regions of interest in the replay buffer viewport.

Vizarel suggests a number of avenues for future research. First, we hypothesize that it could help design metrics that better capture priority during experience replay \cite{schaul_prioritized_2016}. Second, it could help create safety mechanisms early on in the training process through identifying patterns in agent failure conditions \cite{amodei_concrete_2016}. Another possible direction this tool catalyze is the construction of reproducible visualizations through further plugins integrated into the system. 

We anticipate that the best features yet to be built will emerge through iterative feedback, deployment, and usage in the broader reinforcement learning and interpretability research communities.

%===============================================================================

% The maximum paper length is 8 pages excluding references and acknowledgements, and 10 pages including references and acknowledgements

% The acknowledgments are automatically included only in the final version of the paper.
\acknowledgments{SVD is supported by the CMU Argo AI Center for Autonomous Vehicle Research. BE is supported by the Fannie and John Hertz Foundation and the National Science Foundation (DGE1745016). Any opinions, findings, recommendations, and conclusions expressed in this material are those of the author(s) and do not reflect the views of funding agencies.}

%===============================================================================

{\footnotesize
% no \bibliographystyle is required, since the corl style is automatically used.
\bibliography{CoRL}  % .bib
}

\end{document}